\documentclass[sigconf]{aamas} 
\usepackage{balance} % for balancing columns on the final page

\setcopyright{ifaamas}
\acmConference[AAMAS '22]{Proc. Rebellion and Disobedience in AI Workshop, \@ 21st International Conference on Autonomous Agents and Multiagent Systems (AAMAS 2022)}{May 9--13, 2022}
{Auckland, New Zealand}{P.~Faliszewski, V.~Mascardi, C.~Pelachaud,
M.E.~Taylor (eds.)}
\copyrightyear{2022}
\acmYear{2022}
\acmDOI{}
\acmPrice{}
\acmISBN{}

\acmSubmissionID{}

\title[AAMAS-2022 Formatting Instructions]{Rebellion and Disobedience as Useful Tools in Human-Robot Interaction Research --- The Handheld Robotics Case}

\author{Walterio Mayol-Cuevas}
\affiliation{
  \institution{University of Bristol, Amazon.com}
  \country{UK, USA}
}
\email{walterio.mayol-cuevas@bristol.ac.uk}

\begin{abstract}
This position paper\footnote{This position paper is based on the underpinning work with Austin Gregg-Smith and Janis Stolzenwald while at the University of Bristol \cite{GreggSmith:2015bh,GreggSmith:2016cz,GreggSmith:2016hn,Stolzenwald:2018un,Stolzenwald:2019wi}} argues on the utility of rebellion and disobedience (RaD) in human-robot interaction (HRI). In general, we see two main opportunities in the use of controlled and well designed rebellion and disobedience: i) illuminate insight into the effectiveness of the collaboration (or lack of) and ii) prevent mistakes and correct user actions when in the user's own interest. Through the use of a close interaction modality, that of handheld robots, we discuss use cases for utility of rebellion and disobedience that can be applicable to other instances of HRI.
\end{abstract}

\keywords{Human-Robot Interaction, Rebellion in AI, Handheld Robotics}

\newcommand{\BibTeX}{\rm B\kern-.05em{\sc i\kern-.025em b}\kern-.08em\TeX}

\begin{document}

%%% The following commands remove the headers in your paper. For final 
%%% papers, these will be inserted during the pagination process.

\pagestyle{fancy}
\fancyhead{}

%%% The next command prints the information defined in the preamble.

\maketitle 

%%%%%%%%%%%%%%%%%%%%%%%%%%%%%%%%%%%%%%%%%%%%%%%%%%%%%%%%%%%%%%%%%%%%%%%%

\section{Motivation}

Rebellion and disobedience in Human-Robot collaboration may at first attract the notion of a problem to be solved, avoided or even being troubled about. These are all valid concerns, but recent work in rebellion in AI systems has started to look at opportunities and controlled consequences \cite{Aha:Coman:AAAI2017, Coman:Aha:AI2018, Banks:IJSR:2021}.

In this position paper we show how these behaviours can be useful tools in the space of HRI. And in this case, the argument becomes under what circumstances and in what ways it is proper and right for a robot to rebel and or disobey so that human users benefit and tasks have better outcomes.

\begin{figure}[h]
  \centering
  \includegraphics[width=0.24\linewidth]{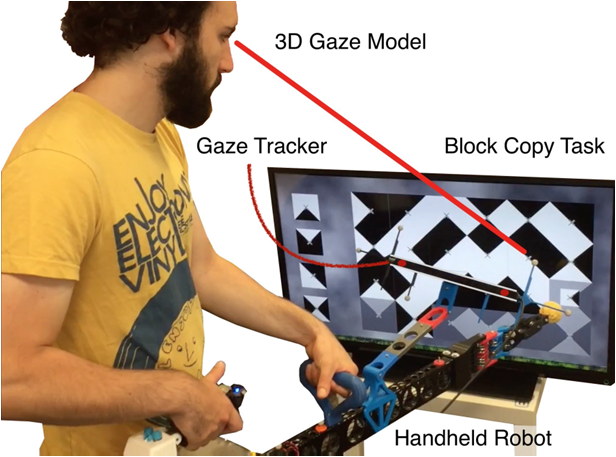}
  \includegraphics[width=0.73\linewidth]{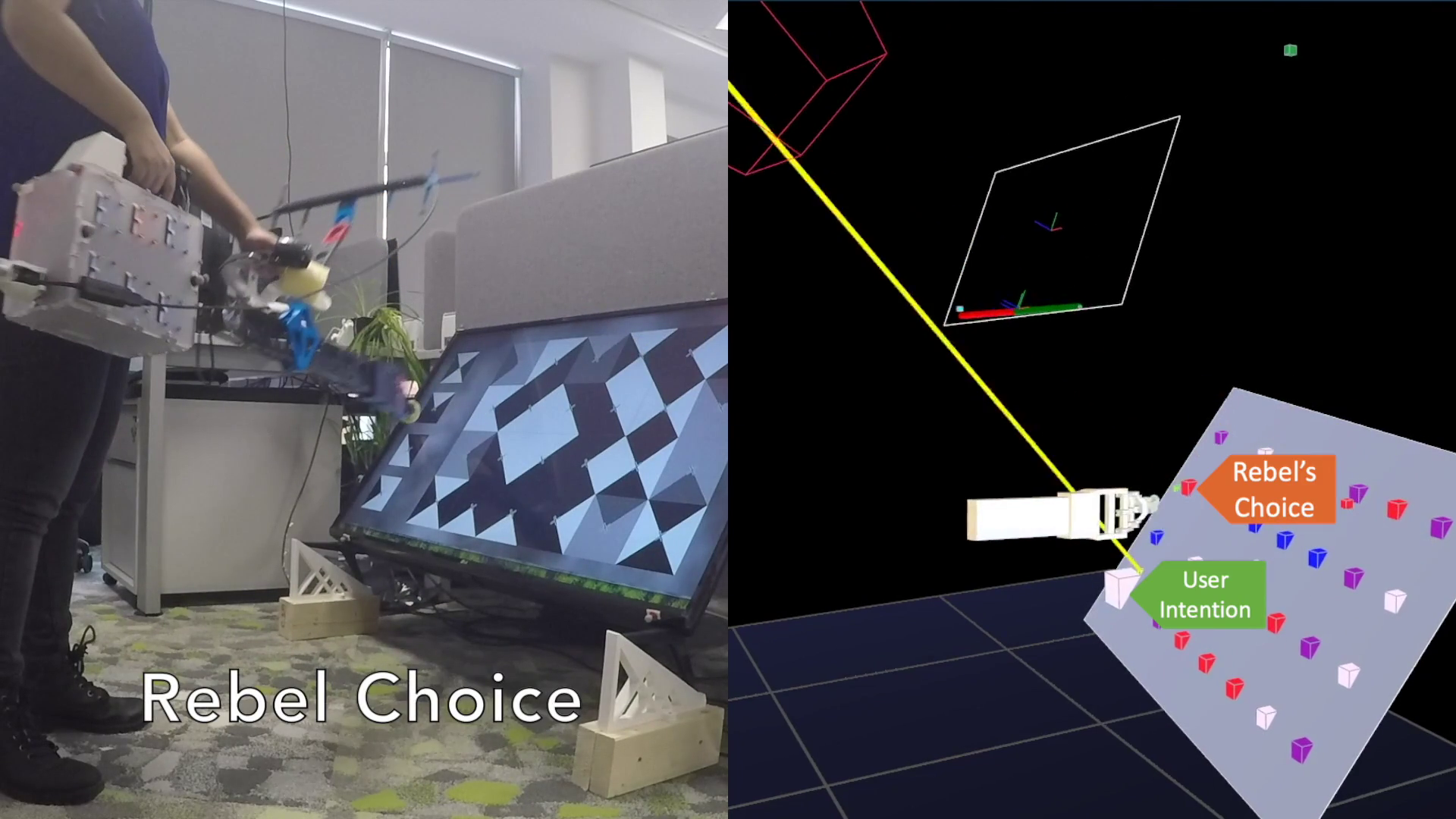}
  \caption{A handheld robot with eye gaze sensing predicts the user's intention (selecting a simulated block in a block-copying task), yet decides to {\it rebel} by placing the block in another valid location \cite{Stolzenwald:2019wi}. This frustrates users but helps researchers understand the level of accuracy on the elusive task of intention prediction. Video at \cite{handheldrobotics}.}
  \label{fig:intention}
%  \Description{}
\vspace{-10mm}
\end{figure}

%%%%%%%%%%%%%%%%%%%%%%%%%%%%%%%%%%%%%%%%%%%%%%%%%%%%%%%%%%%%%%%%%%%%%%%%

\section{HRI with Handheld Robotics}

Handheld robots \cite{GreggSmith:2015bh,GreggSmith:2016cz,GreggSmith:2016hn,Stolzenwald:2018un,Stolzenwald:2019wi, handheldrobotics, Elsdon:2017is,Elsdon:2018kl} are a distinct type of robot that interacts closely with users. A handheld robot has the overall usability and form of handheld tools but with added sensing, actuation, and importantly, task knowledge.

Handheld Robots can be considered as intelligent tools that can process task knowledge and environment information for semi-autonomous or shared assistance in collaborative task solving. Handheld robots combine the robotic abilities of speed, force, extended memory and accuracy with the natural competences of human users e.g. negotiating obstacles and resolving complex motion-planning tasks effortlessly. 

Through their features and use, handheld robots marry the two ends of Moravec's paradox \cite{Moravec:1988un}, which states that actions that are hard for machines are easy to complete for humans and vice versa. However, in handheld robotics, and due to the close proximity to users, this paradox becomes a collaborative symbiosis: users carry out the tactical motions following intuitive navigation and obstacle avoidance skills and benefit from the robot's super-human performance in speed, accuracy and task knowledge. The result is a potential reduction in the time needed for task execution and in the number of errors. This collaboration helps users of handheld robots to carry out tasks in which they may have limited skill or expertise. 

In this context, handheld robots are an interesting space to study the effects of rebellion and obedience. In particular by being close, albeit not {\it attached} to the user, if a handheld rebels or disobeys commands it is important to design the right framework to communicate between user and robot. Or better yet, if the disagreement is for the benefit of the user's task progression, how to graciously decline the execution of commands and indicate or even carry out the correct actions with reduced frustration. This dynamic shifting of expertise can also lead to teaching and up-skill opportunities for both robot and user respectively.

Handheld robots can find applications in the areas where current handheld tools operate. These include construction, agriculture, maintenance, recreation, the arts, among others.

%\begin{figure}[h]
%  \centering
%  \includegraphics[width=0.45\linewidth]{MK1.png}
%  \caption{Caption}
%  \label{fig:MK1}
%  \Description{Description}
%\end{figure}

%%%%%%%%%%%%%%%%%%%%%%%%%%%%%%%%%%%%%%%%%%%%%%%%%%%%%%%%%%%%%%%%%%%%%%%%

\section{The use of Rebellion and Disobedience in HRI research}

The general argument in this position paper is that rebellion and disobedience are useful in not only the study of HRI in general, but that these behaviours can be helpful to gain greater insight into the collaboration between human and robot, and as a necessary tool for safety and performance.

To this end, here we describe two ways by which rebellion and disobedience have been used in our handheld robot research.

\subsection{Disobedience for task performance}
The most natural way to expect a beneficial introduction of disobedience in HRI is to intervene in cases and moments when the user is about to make a mistake or if the expected performance will be outside of acceptable parameters or a risk to the user.
Consider a smart driller that is about to drill through a pipe that is known to the robot through floor plans or even known to the user but due to inexperience or a cognitive block, the user is unable to stop. Such a smart driller would be able to prevent the action before damage is caused, or at least help mitigate it. In effect, the driller would have disobeyed a command but in this instance for a greater good according to the task and available information.

\begin{figure}[h]
  \centering
  \includegraphics[width=0.2\linewidth]{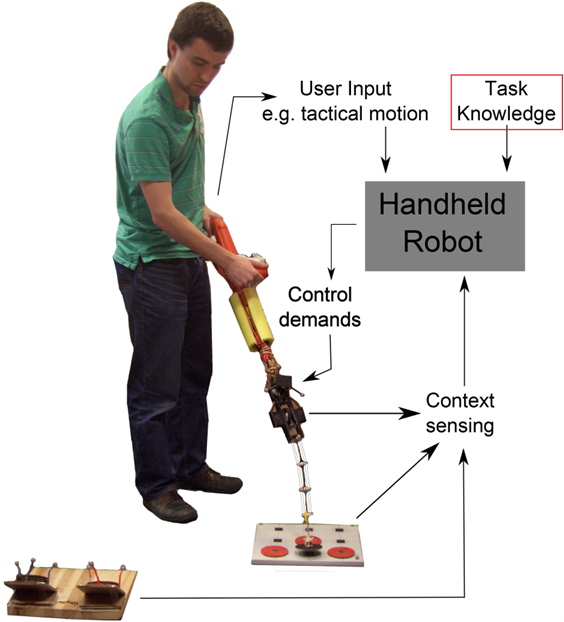}
    \includegraphics[width=0.2\linewidth]{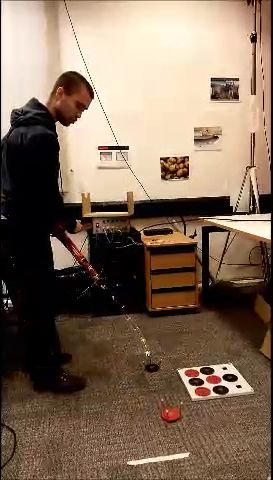}
    \includegraphics[width=0.2\linewidth]{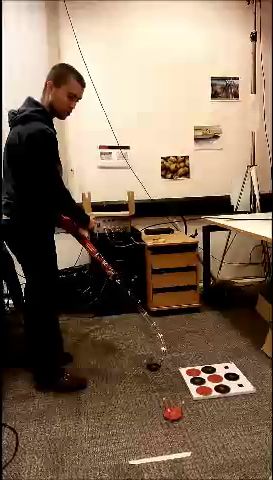}
    \includegraphics[width=0.2\linewidth]{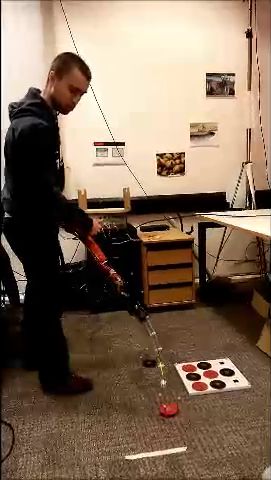}
  \caption{A handheld robot that has task knowledge refuses to perform an action that is incorrect and guides the user to the correct action via gestures \cite{GreggSmith:2015bh}. Video at \cite{handheldrobotics}.}
  \label{fig:Disobedience}
%  \Description{Description}
\vspace{-3mm}
\end{figure}

In figure \ref{fig:Disobedience}, a handheld robot \cite{Stolzenwald:2019wi} is tasked to cooperate with the user to place coloured tiles (Red and Black) according to a pattern. The user tactically moves the robot from one tile stack to the tiling board and back. In this case the user has freedom to select the sequence of tiles and order. This allows the user to have a strong sense of agency which is important in cooperative robotics. The robot in turn is helping by placing tiles in a precise location as well as monitoring task progression.

Eventually, the user is incorrectly trying to pick up an additional black tile. But since the robot has been monitoring the task, it is aware that no further black tiles are needed. In this case the robot {\it refuses} to pick up more black tiles and by simply pointing to the red stack it uses gentle gestures to indicate the next correct step. This is an example of graciously declining an incorrect action and nudging the user towards the correct one. Gesturing is a powerful, intuitive way to convey spatial information that can rival screens \cite{GreggSmith:2016hn}. Spatial pointing gestures can also be easier to localise vs text or vocal instructions, expanding their universality. 

%\begin{figure}[h]
%  \centering
%  \includegraphics[width=0.95\linewidth]{Drawing Performance ICRA2015.png}
%  \caption{Caption}
%  \label{fig:MK1}
%  \Description{Description}
%\end{figure}

\subsection{Rebellion to surface prediction accuracy}

While preventing a mistake is a clear reason for an HRI system to disobey, it is also useful to consider how disobedience and rebellion can be used in other ways. In this case the system in figure \ref{fig:intention}, the handheld robot has the ability to predict user intention from the combination of gaze sensing (yellow ray), and task knowledge \cite{Stolzenwald:2019wi}. In this case the task is a block copying task where users must take virtual tiles from the source location and place them in appropriate locations on the tiling area. The shape of the tile needs matching with the intended target location and users need to select tiles and place them according to the expected pattern.

Users were able to freely decide the starting order of placements to encourage agency. However, a system that is intended to anticipate what users want to do is one step further towards sophisticated helping in HRI. In this scenario once the robot estimates what the user wants, it can accelerate the sub-task completion by rapidly moving towards where the user intends to place the tile and therefore speed-up task completion. The system predicts based on task knowledge combined with gaze fixations, which in humans reveal intention \cite{Land:2001hl}. 

In this instance, the robot rebels on purpose, by first predicting where the user wants to place the next tile and instead of helping, it avoids the predicted placing location and chooses an alternative location. The alternative location is a valid location for that specific tile shape so that the task is seemingly unaffected, yet the user ends being frustrated {\it intentionally}, as the location the robot uses is not the user's intended one.

\begin{figure}[h]
		\vspace{-1.8em}
		\centering
		\includegraphics[width=0.75\linewidth]{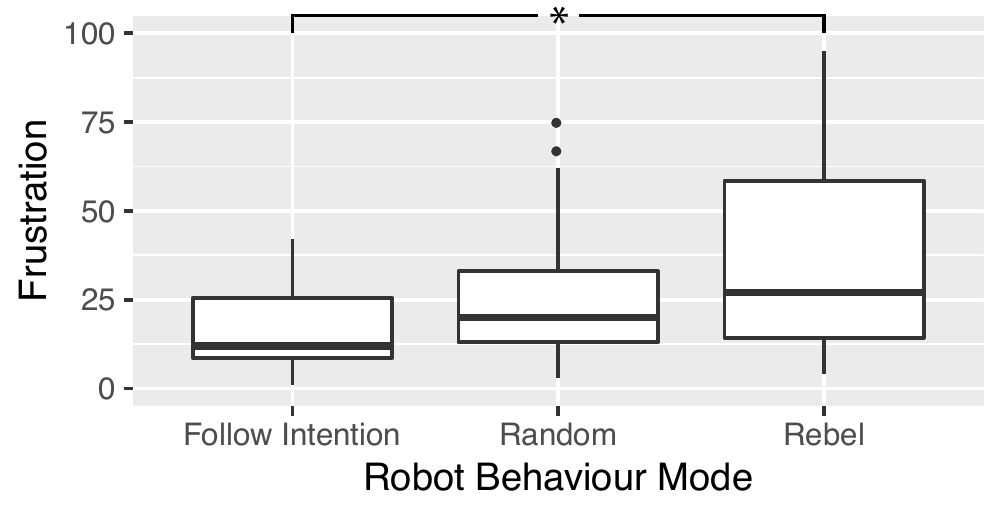}
%		\vspace{-0.6em}
		\caption{Perceived frustration via TLX survey for each of the tested behaviour modes. Frustrating users on purpose helps to gain confidence on intention prediction for better helping users. * indicates significant difference \cite{Stolzenwald:2019wi}. }
		\label{fig:intentionvalidationfrustration}
		\vspace{-3mm}
	\end{figure}

Predicting intention is one of the hardest task in HRI, in part because the intention signal is not easy or desirable to explicitly state. Therefore, in this case, rebellion which results in user frustration, is used as an enhancer to the signal on intention prediction --- if the user is frustrated from purposeful rebellion, the confidence that the system is correctly predicting is greater when taken together with the signal from when the user does follow predictive intention. This form of rebellion or contrarian approach we hypothesise is potentially also useful to reduce the influence of novelty in HRI. Frustration is harder to hide on user questionnaires and helps to mitigate volunteers' potential bias to guess what the researcher's intended outcome is. Figure \ref{fig:intentionvalidationfrustration} shows the difference, and thus our hypothesised improved confidence between signals when the robot predicts and helps and predicts and rebels.

%\begin{figure}[h]
%  \centering
%  \includegraphics[width=0.45\linewidth]{IntentionPrediction.png}
%  \caption{Caption}
%  \label{fig:intention}
%  \Description{Description}
%\end{figure}
`

\subsection{Conclusions and Discussion}
The use of rebellion and disobedience in HRI is an emerging area. We argue for two cases where the inclusion of this behaviour is useful. The first being the more expected one that is to prevent mistakes or danger, but there is also the utility of rebellion to better surface signals of collaboration such as intention prediction. We also believe there are further uses of RaD in Robotics and AI and deeper understanding of rebellious robot behaviour and the ways to handle it will benefit HRI at large.

%%%%%%%%%%%%%%%%%%%%%%%%%%%%%%%%%%%%%%%%%%%%%%%%%%%%%%%%%%%%%%%%%%%%%%%%

%%% The acknowledgments section is defined using the "acks" environment
%%% (rather than an unnumbered section). The use of this environment 
%%% ensures the proper identification of the section in the article 
%%% metadata as well as the consistent spelling of the heading.

%\begin{acks}
%If you wish to include any acknowledgments in your paper (e.g., to 
%people or funding agencies), please do so using the `\texttt{acks}' 
%environment. Note that the text of your acknowledgments will be omitted
%if you compile your document with the `\texttt{anonymous}' option.
%\end{acks}

%%%%%%%%%%%%%%%%%%%%%%%%%%%%%%%%%%%%%%%%%%%%%%%%%%%%%%%%%%%%%%%%%%%%%%%%

%%% The next two lines define, first, the bibliography style to be 
%%% applied, and, second, the bibliography file to be used.

\bibliographystyle{ACM-Reference-Format} 
\bibliography{sample}

%%%%%%%%%%%%%%%%%%%%%%%%%%%%%%%%%%%%%%%%%%%%%%%%%%%%%%%%%%%%%%%%%%%%%%%%

\end{document}